\documentclass{article}

\usepackage{longcat_style}
\usepackage{adjustbox}
\usepackage[utf8]{inputenc} %
\usepackage[T1]{fontenc}    %
\usepackage{newunicodechar}
\newunicodechar{♫}{\raisebox{-0.2em}{\includegraphics[height=1.2em]{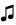}}}
\newunicodechar{‐}{-} 
\usepackage{hyperref}       %
\usepackage{xcolor}
\usepackage[normalem]{ulem} %
\hypersetup{
    colorlinks=true,      %
    linkcolor=blue,      %
    urlcolor=blue,       %
    citecolor=blue,      %
    linkbordercolor=blue, %
    urlbordercolor=blue,
    citebordercolor=blue,
    pdfborderstyle={/S/U/W 1}, %
}

\usepackage{float}
\usepackage{wrapfig}
\usepackage{placeins}
\usepackage{url}            %
\usepackage{booktabs}       %
\usepackage{array}
\usepackage{amsfonts}       %
\usepackage{nicefrac}       %
\usepackage{microtype}      %
\usepackage{lipsum}		%
\usepackage{graphicx}
\usepackage{natbib}
\usepackage{doi}
\usepackage{amsmath}
\usepackage{amssymb} %
\usepackage{xspace}
\usepackage{enumitem}
\usepackage{multirow}
\usepackage{subcaption} 
\usepackage{makecell}
\usepackage{hyperref}
\usepackage[capitalise]{cleveref}
\usepackage{pifont}
\usepackage[inkscapelatex=false]{svg}
\usepackage{subcaption}
\usepackage[most]{tcolorbox} %
\usepackage{pgfplots}
\pgfplotsset{compat=1.18}
\usepgfplotslibrary{groupplots}

\newtcolorbox{summarybox}{
    colback=blue!5!white,    %
    colframe=black!75,       %
    arc=6pt,                 %
    boxrule=1pt,             %
    left=8pt,                %
    right=8pt,               %
    top=8pt,                 %
    bottom=8pt,              %
    enhanced,                %
}

\setlist[itemize]{leftmargin=*}
\setlist[enumerate]{leftmargin=*}
\setlist[description]{leftmargin=*}

\definecolor{midnightgreen}{rgb}{0.0, 0.29, 0.33}

\title{LongCat Recall Sparse Attention: Co-Designing Key Organization and Retrieval for Long-Context Inference}

\author{\large Hao Zhang\textsuperscript{\normalfont *}\qquad
\large Bin Xiao\textsuperscript{\normalfont *,$\dagger$}\qquad
\large Kefeng Zhang\\[3pt]
{\normalfont\normalsize Meituan LongCat Team}\\[4pt]
{\normalfont\small \textsuperscript{*}\,Equal contribution.\qquad
\textsuperscript{$\dagger$}\,Project lead and corresponding author.}
}

\renewcommand{\headeright}{\raisebox{-0.2\height}{\includegraphics[height=1.8em]{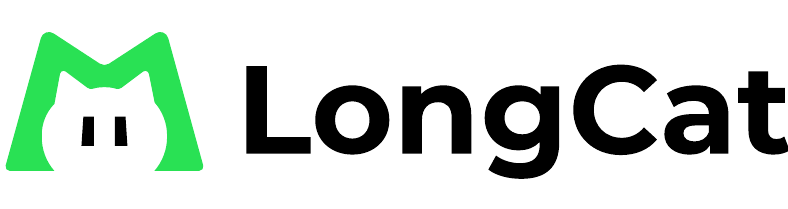}}}
\renewcommand{\shorttitle}{LongCat Recall Sparse Attention}

\begin{document}
\raggedbottom
\setlength{\headheight}{17pt}
\addtolength{\topmargin}{-3pt}
\maketitle

\begingroup
\renewenvironment{abstract}
{\centerline{\large\bfseries\scshape Abstract}\smallskip\par\small\noindent}
{\par}
\begin{abstract}
Retrieval-based sparse attention, exemplified by DeepSeek Sparse Attention
(DSA), reduces attention computation by selecting a small subset of
key--value entries. However, selecting these entries can still require
searching the entire visible history. For long sequences, retrieval becomes
a major bottleneck, limiting the efficiency gains from sparse attention.
To address this problem, we propose LongCat Recall Sparse Attention (LRSA), which combines fine-grained
representatives with paged retrieval in an incrementally maintained index.
LRSA partitions the sequence into time buckets that can be indexed
independently as new keys arrive, without regrouping existing history.
Within each bucket, it organizes indexing keys into capacity-balanced
directional groups and
packs a fixed number of complete groups into each page, retaining a
separate center for every group. These centers provide fine-grained
representatives for page selection, after which refinement scores members
of the selected pages together with sink, sliding-window, and visible
unindexed candidates to select the final Top-$K$ logical positions.
To execute this retrieval process efficiently, we store members in
reordered pages, fuse construction stages according to group size, and
share storage for overlapping fixed candidates across queries during
chunked prefill.
These choices support regular member access while reducing kernel launches
during construction and repeated candidate transfers. Moreover, we integrate LRSA into
SGLang to maintain the index during inference, overlapping asynchronous
index construction for new buckets with subsequent computation and reusing compatible bucket
indices and member pages after a prefix match in the original key--value
pool.
With a 1024K-token context, LRSA accelerates key selection by 4.70$\times$ over DSA and combined selection and attention by 35.60$\times$ over full attention.

\end{abstract}
\endgroup

\begin{figure}[!ht]
    \centering
    \captionsetup{font=small}
    \includegraphics[width=\linewidth]{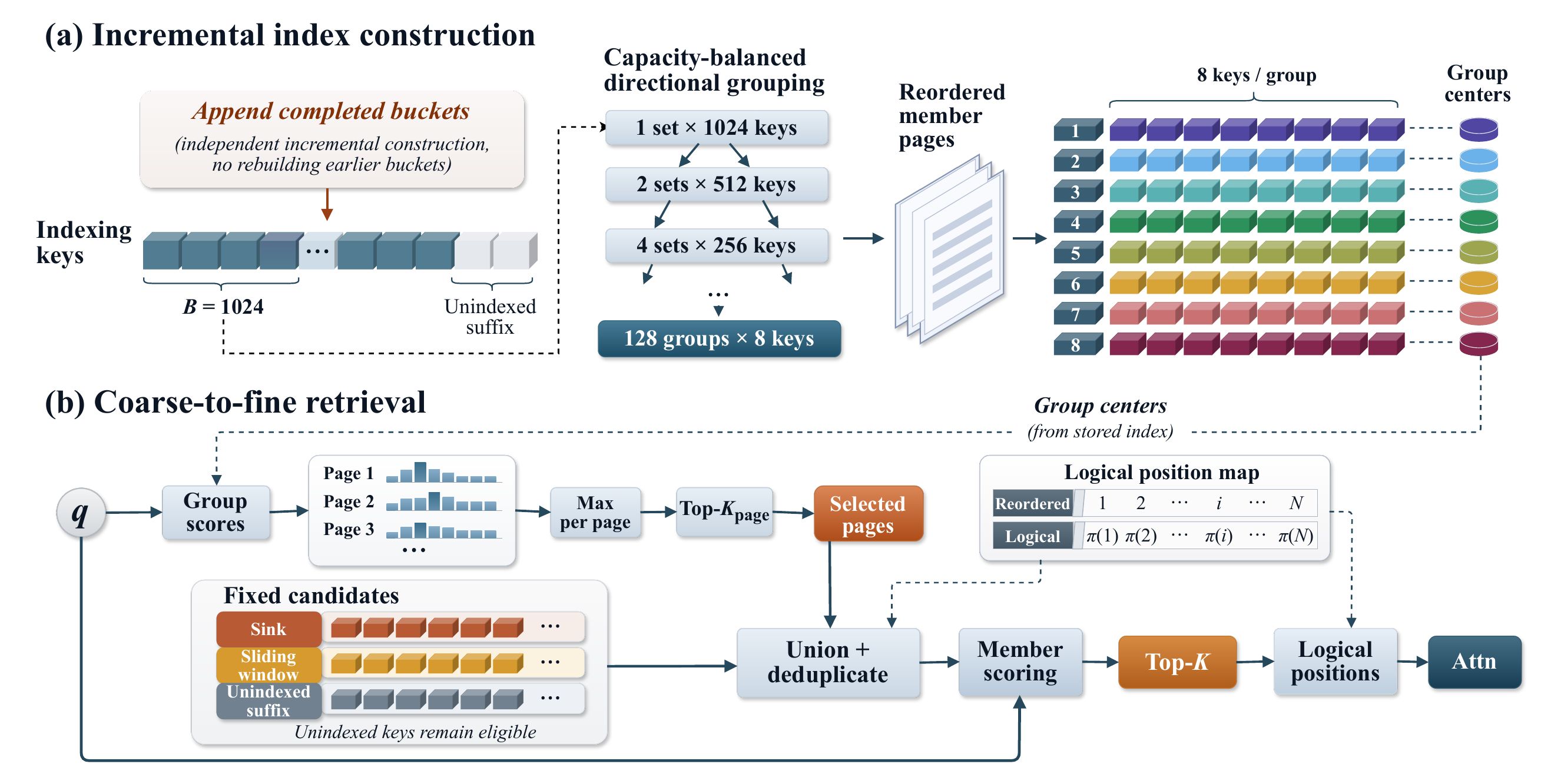}
    \caption{Overview of LRSA. (a) Each completed bucket is independently
    partitioned into capacity-balanced directional groups, which are packed
    into member pages. With the default capacities, the expanded page
    contains eight groups of eight keys, with one center per group.
    (b) Queries score group centers and select the Top-$K_{\mathrm{page}}$
    pages using the maximum group score within each page. Retrieved members are
    combined with sink, sliding-window, and visible unindexed candidates,
    deduplicated, and scored to select the final Top-$K$ positions for
    attention. The position map supports deduplication and recovery of
    original logical positions. Reordered page and slot addresses are
    depicted as one-dimensional indices for simplicity.}
    \label{fig:teaser}
\end{figure}

\clearpage
\tableofcontents
\clearpage

\section{Introduction}
\label{sec:intro}
Long-context processing addresses the need for large language models
(LLMs) to draw on extensive documents, code, and interaction histories
in applications such as book-length summarization
\citep{wu2021books}, repository-level code completion
\citep{zhang2023repocoder}, and memory-based interactive agents
\citep{park2023agents}. However, dense attention attends to all causally
visible positions for each query, so the total computation over a sequence
grows quadratically with its length.
Retrieval-based sparse attention, exemplified by DeepSeek Sparse Attention
(DSA) \citep{deepseek2025v32}, reduces this cost by identifying relevant
key--value (KV) entries and restricting the main attention computation to
the selected subset. DSA performs this selection with a
lightweight learned indexer that scores visible keys and returns their
Top-$K$ positions. Although a fixed selection budget limits the number of
entries used by the main attention, selecting these entries still requires
the indexer to score every visible key. The cost of this exhaustive search
grows with context length. For long sequences, retrieval becomes a major
bottleneck, limiting the
efficiency gains from sparse attention. Therefore, reducing the cost of key selection in sparse attention is
central to efficient long-context inference.

Recent work reduces key-selection costs through complementary approaches.
HISA \citep{xu2026hisa} uses pooled representations of contiguous blocks
to narrow the search before scoring individual keys, while IndexCache
\citep{bai2026indexcache} reuses selected indices across layers. LiteTopK
\citep{yin2026litetopk} fuses scoring with exact Top-$K$ selection to reduce
selection overhead. LongCat Sparse Attention \citep{zan2026longcat}
combines hierarchical retrieval with cross-layer index reuse and
streaming-aware budget allocation. These approaches highlight the need
to consider both the search procedure and its execution. In this work, we focus on how
historical keys are organized for retrieval, since this organization shapes
both how well group representatives capture relevant keys and how
efficiently group members can be accessed.

This organization must reconcile several requirements. Directional
grouping, as used by ClusterKV \citep{liu2024clusterkv}, provides a basis
for coherent group representatives. Unconstrained group sizes, however,
complicate the mapping of groups to fixed-capacity pages: groups may span page
boundaries, and using them as retrieval units produces uneven refinement
workloads across groups. Capacity-constrained groups can instead
be packed as complete units into pages, giving each page a regular group
structure for member access. Retaining a separate center for each small
group also avoids obscuring a relevant subset within a single page-wide
mean. The index must also grow as new keys arrive without repeatedly
regrouping existing history, while keeping visible unindexed keys eligible
for selection.
These requirements suggest separating the granularity of representation
from that of access, while giving construction its own fixed update units.

To meet these requirements, we propose \emph{LongCat Recall Sparse
Attention (LRSA)}, which organizes historical keys into a hierarchy of
time buckets, directional groups, and retrieval pages. Each bucket can be
indexed independently, without regrouping keys in existing buckets. Within
each bucket, recursive directional
grouping produces capacity-balanced groups, and each page packs a fixed
number of complete groups while retaining a separate center for each.
A query scores these centers and selects pages using the maximum center
score within each page, allowing a relevant subgroup to influence
retrieval without averaging its score with those of other groups.
Refinement then combines the retrieved members with fixed candidates
from the sink prefix, sliding window, and visible unindexed suffix, and applies
the indexer's scoring function to select the final Top-$K$ logical
positions. The page budget bounds the number of historical members retrieved through the index, while the fixed
candidates remain eligible regardless of which pages are selected.

This page structure also provides the basis for efficient execution.
We materialize the groups as reordered member pages for contiguous access,
with a mapping to original logical positions for duplicate handling and
subsequent attention. To construct these pages, recursive
partitioning updates compact member indices instead of repeatedly moving
full vectors, and group-size-dependent kernel fusion reduces launches and
intermediate memory traffic. For refinement, we copy the fixed candidates
into the reordered pool so that they can be scored together with retrieved
members through the same paged interface. Because their ranges overlap
across neighboring queries during chunked prefill, we store their union
once per chunk, avoiding
repeated transfers of the same keys while preserving each query's candidate
set.

Maintaining this index as inference proceeds requires coordinating its
updates with model execution and managing the lifetime of its resources
within and across requests. We integrate LRSA into SGLang, where indices for newly filled
buckets are constructed asynchronously after the current layer's attention
so that construction can overlap with subsequent computation. The bucket and page structure
also supports reuse across requests. After a prefix match in the original
KV pool, valid and compatible indices for buckets fully covered by the
matched prefix can be reused to skip construction; when a bucket within
that prefix must be rebuilt, matching
member pages can be shared after assembly to avoid duplicate storage.
Retention policies govern how long these resources remain available for
reuse, trading index memory occupancy against reconstruction cost.
Experiments show that, at a context length of 1024K tokens, LRSA achieves a 4.70$\times$ speedup in key selection over DSA and a 35.60$\times$ speedup in combined key selection and attention computation over full attention.

Our main contributions can be summarized as follows:
\begin{itemize}
    \item We propose a retrieval hierarchy that combines fine-grained
    directional representatives with fixed-capacity pages through
    capacity-constrained grouping, supporting regular member access and
    predictable historical refinement workloads. Time buckets enable
    independent index updates without regrouping existing keys.
    \item We develop a kernel implementation that reduces kernel launches
    during index construction and repeated candidate transfers through
    group-size-dependent fusion and shared candidate storage for chunked
    prefill.
    \item We integrate LRSA into SGLang, coordinating asynchronous index
    construction, prefix-based reuse, and resource retention throughout
    the index lifecycle.
\end{itemize}

\section{Retrieval-Based Sparse Attention and Design Motivation}
\label{sec:related}

We review DeepSeek Sparse Attention and related approaches to reducing
retrieval overhead, then motivate our joint design of key organization,
paged access, and incremental maintenance.

\subsection{DeepSeek Sparse Attention and Subsequent Developments}
\label{sec:dsa-background}

Retrieval-based sparse attention first identifies relevant historical positions
and computes attention over the corresponding KV entries. DeepSeek Sparse
Attention (DSA) \citep{deepseek2025v32} implements this approach with a learned
\emph{lightning indexer}. For a query at position $t$, the indexer assigns each
causally visible indexing key $k_i^{I}$ the score
\begin{equation}
    s_t(i) = \sum_{h=1}^{H_I} w_{t,h}\,
    \operatorname{ReLU}\!\left((q_{t,h}^{I})^{\top} k_i^{I}\right),
    \qquad i \leq t,
    \label{eq:dsa-score}
\end{equation}
where $H_I$ is the number of indexing heads, $q_{t,h}^{I}$ is the projected
query for head $h$, and $w_{t,h}$ is its query-dependent weight. The $K$
highest-scoring positions determine the KV entries accessed by the main
attention. The indexer uses lightweight, FP8-compatible scoring and is
supervised by the main attention distribution through a KL-divergence loss.
Its scores guide selection, while the main attention computes its own scores
and normalization over the retrieved entries. For a sequence of length $L$
and a fixed selection budget $K$, the main attention requires $O(LK)$ work.
However, scoring every visible key for every query retains an $O(L^2)$
indexing cost.

Recent work reduces retrieval costs through search reduction, cross-layer
reuse, and execution optimizations. These directions address different
sources of overhead and can be combined.

\emph{Search reduction.} HISA \citep{xu2026hisa} pools indexing keys in
contiguous blocks, selects blocks using their representatives, and scores
individual keys within the selected blocks for refinement. MISA \citep{zhou2026misa}
activates a subset of indexing heads for each query, optionally refining
the resulting candidates with all heads. SPIN \citep{fu2026spin} predicts
block importance from historical indexer scores and evaluates the current
query only against selected blocks, using exploration to refresh the
statistics of skipped blocks.

\emph{Cross-layer reuse.} IndexCache \citep{bai2026indexcache} reuses final
Top-$K$ indices across layers. MLIPS \citep{jain2026mlips} instead shares a
larger candidate pool from which each consumer layer independently selects
its entries. Prism Attention \citep{tian2026prism} similarly transmits
expanded proposals from designated layers and refines them using the
receiving layer's scores; quantized keys reduce proposal-generation cost.
These approaches amortize retrieval across layers while differing in how
much layer-specific selection they retain.

\emph{Execution efficiency.} LiteTopK \citep{yin2026litetopk} fuses scoring
with exact Top-$K$ selection, and LiteDSA shares
selected KV loads across neighboring queries using per-query membership
masks. DoSA \citep{meng2026dosa} uses score bounds to avoid full evaluation
of some candidates while preserving exact Top-$K$ selection under its
stated conditions. JSSA \citep{song2026jssa} combines a joint low-dimensional
query--key scoring subspace with a gather-free indexed-attention kernel.
LongCat Sparse Attention \citep{zan2026longcat} combines hierarchical
retrieval, cross-layer reuse with distillation, and streaming-aware
indexing, bringing together search reduction, reuse, and locality
optimizations.

\emph{Positioning LRSA.} LRSA focuses on how the organization of keys
connects representative-based retrieval to paged execution. HISA pools
keys within contiguous blocks, while LongCat Sparse Attention retains
a mean vector for each contiguous sub-block within a page and sums the
resulting sub-block scores. LRSA instead forms capacity-balanced
directional groups within each time bucket and packs
complete groups into reordered pages. It scores each page by its
highest-scoring group, so the strongest group-level match determines the
page's retrieval priority. This couples fine-grained directional representatives with
fixed-capacity access units: representation follows key similarity, while
member retrieval follows a regular paged layout.

Semantic grouping is also central to ClusterKV \citep{liu2024clusterkv},
which ranks clusters and accumulates their sizes to meet a token budget,
trimming the last cluster when needed. LRSA imposes capacity constraints
during grouping, then separates page retrieval from member-level Top-$K$
selection. Group capacity determines representation granularity, page
capacity determines access granularity, and member-level scoring selects
the final entries under a separate attention budget. Small groups can
therefore guide retrieval while their members are accessed in fixed-capacity
pages.

Beyond key organization, LRSA differs in how it obtains and prepares
candidates. SPIN predicts block importance from historical
scores, whereas LRSA evaluates group centers with the current query.
IndexCache, MLIPS, and Prism Attention reuse indices or proposals across
layers; LRSA searches an index maintained within each layer. JSSA changes
the scoring representation and final KV access, while LRSA narrows the
candidate set through approximate page retrieval and retains the indexing
function in Eq.~\eqref{eq:dsa-score} for member-level refinement.

At the execution level, LiteDSA shares selected KV loads during attention,
whereas LRSA shares the indexing keys of fixed candidates before refinement.
These candidates comprise sink, sliding-window, and visible unindexed
positions. Shared storage avoids repeatedly copying their overlapping
ranges for neighboring queries. To maintain the retrieval organization as
new keys arrive, LRSA builds each time bucket independently, preserving the
groups in previously indexed buckets. Together, these choices connect
directional representation and fixed-capacity page access to candidate
preparation and incremental maintenance.

\subsection{Motivating the Design of Efficient Sparse Retrieval}
\label{sec:motivation}

Our design addresses three related aspects of retrieval efficiency: reducing
the history scanned by each query, organizing keys into groups that support
informative representatives and align with paged access, and managing index updates
and candidate preparation as the context grows. The following
considerations explain why these aspects must be addressed together.

\emph{Full-history scoring remains a bottleneck under a fixed attention budget.}
With $H_I$ indexing heads of dimension $d_I$, scoring $n$ visible keys for
one query requires $O(H_I d_I n)$ work. For $R$ queries with visible lengths
$n_q$, the total scoring cost is $O(H_I d_I \sum_{q=1}^{R} n_q)$, with
additional work to select the Top-$K$ entries for each query. Over a causal prefix of
length $L$, $\sum_{q=1}^{L} q = L(L+1)/2$, so exhaustive indexing incurs
quadratic scoring work even when the final attention budget is fixed.
This cost motivates a hierarchical search: score a smaller set of group
representatives first, then refine the search by scoring individual keys
within the selected pages.

\emph{Grouping must support informative representatives and regular paged access.}
Retrieval quality depends on how well each representative captures the
keys in its group. Temporal proximity alone does not ensure directional
similarity: the mean of keys with different directions may poorly represent
a member that scores highly for a given query. Grouping by directional similarity,
as in ClusterKV, provides a basis for more coherent
representations. However, without capacity constraints, groups can differ
substantially in size. Using these groups as retrieval units produces
uneven refinement workloads across groups, and packing them into
fixed-capacity pages may split groups across page boundaries and complicate
the mapping between centers and member pages. We therefore combine
directional grouping with capacity constraints so that each page contains
a fixed number of complete groups. This preserves small-group
representatives while supporting regular member access. Fixed page
capacity then translates a page-selection budget into a bound on the
number of retrieved historical members and the associated member-level
scoring work.

\emph{Maintenance and candidate preparation determine the net benefit of search.}
The organization used for retrieval must also accommodate newly arriving
keys. Rebuilding the full index repeats work on historical data, while
searching only the existing index leaves visible but unindexed keys outside
the candidate set. We therefore extend the index incrementally and make
completed buckets available for retrieval. Visible keys awaiting
indexing are included directly as refinement candidates, preserving their
eligibility for selection during index construction. Candidate preparation
also incurs data movement: neighboring queries often require overlapping
fixed prefix and tail ranges, so copying them separately for each query
transfers the same data repeatedly. Sharing storage for these candidates,
while retaining each query's visibility constraints, avoids this duplication.
Search, index maintenance, and candidate preparation must therefore be
considered together: reduced
historical scanning yields a net benefit only when its savings exceed the
additional costs of organizing and maintaining retrieval.

\section{LongCat Recall Sparse Attention (LRSA)}
\label{sec:method}

LRSA reduces historical scanning through a hierarchy of time buckets,
directional groups, and retrieval pages. Time buckets delimit independent
index updates. Within each bucket, small groups provide representative
centers, while pages combine several groups into units for coarse retrieval.
A query uses group centers to select pages, then scores their member keys
together with keys from the sink, sliding window, and unindexed suffix.
This separates the organization of history from query-dependent selection,
allowing the index to grow without repeatedly regrouping existing keys.
We first describe how buckets are maintained during inference,
then explain their internal construction and the two-stage retrieval process.

\subsection{Temporal Bucketing and Incremental Index Maintenance}
\label{sec:lrsa-overview}

For each attention layer, we partition indexing positions in temporal order
into consecutive buckets of capacity $B$:
\begin{equation}
    \mathcal{B}_m=\{mB+1,\ldots,(m+1)B\}, \qquad m=0,1,\ldots.
    \label{eq:lrsa-buckets}
\end{equation}
A bucket is the unit of index construction and publication, rather than the
unit selected by a query. Directional grouping is confined to each bucket,
and the resulting groups and centers remain unchanged after construction.
As new keys arrive, only newly filled buckets require index construction.
The bucket boundaries stay fixed as the context grows, while sink and sliding-window
membership are handled during retrieval. Moving the window does not require
rebuilding the historical index.

We set the prefill chunk size equal to the bucket capacity, with $B=1024$
by default. A chunk is a unit of model execution, whereas its corresponding
bucket determines which keys are indexed together. Each full prefill chunk
supplies one bucket per layer. Its queries use the previously published
index, and the current bucket is submitted for construction after that
layer's attention computation. Decoding uses the same granularity: arriving
positions gradually fill the next bucket, and construction is triggered
when that bucket contains $B$ positions. Any partially filled bucket left by prefill is
completed by subsequent arrivals. Thus both stages update the index in
increments of $B$, with prefill supplying these positions together and
decoding accumulating them over time.

A full bucket becomes queryable only after construction has finished and
its records have been published. Let $E$ denote the last position in the
contiguous published prefix. The centers, member pages, and position mapping
for positions $1,\ldots,E$ form a consistent index snapshot. Every visible
position beyond $E$ remains a direct refinement candidate, including
positions in an incomplete bucket. Once the next bucket is completed and
added to the index, the searchable prefix extends by $B$ positions, and
$E$ is updated to $E+B$. This separates data visibility from index readiness:
a new key can participate in selection before its bucket becomes searchable.

We launch index construction asynchronously after the current layer's
attention, allowing it to overlap with subsequent layers and other
independent computation. This schedule aims to complete construction before
the next query batch reaches the same layer, hiding the index-building
latency behind ongoing model computation.

\subsection{Capacity-Balanced Grouping and Page Construction}
\label{sec:lrsa-build}

Each full bucket is organized into groups of capacity $C_{\mathrm{grp}}$,
which are packed into pages of capacity $C_{\mathrm{page}}$. We require
$C_{\mathrm{grp}}$ to divide $C_{\mathrm{page}}$ and $C_{\mathrm{page}}$ to
divide $B$. With the defaults $C_{\mathrm{grp}}=8$ and
$C_{\mathrm{page}}=64$, each page contains
$N_{\mathrm{grp}}=C_{\mathrm{page}}/C_{\mathrm{grp}}=8$ groups.
A bucket of $B=1024$ positions consequently produces 128 groups and 16 pages.
Groups determine the resolution of the representation, while pages determine
the granularity of retrieval and member access. The divisibility conditions
allow each page to contain a fixed number of complete groups, with no
group spanning member-page boundaries.

To construct the groups, we normalize each indexing key for directional
comparisons:
\begin{equation}
    u_i=\frac{k_i^{I}}{\max(\lVert k_i^{I}\rVert_2,\epsilon)},
    \label{eq:lrsa-directions}
\end{equation}
where $\epsilon>0$ protects against zero norms. We initialize the member
list $I$ in the bucket's temporal order and recursively partition it.
At each split, we use the first member $v$ of the current list as the initial
reference and select two directional anchors,
indexed by $i_{-}$ and $i_{+}$, through successive minimum-similarity searches:
\begin{equation}
    i_{-}=\operatorname*{arg\,min}_{i\in I}u_i^{\top}u_v,
    \qquad
    i_{+}=\operatorname*{arg\,min}_{i\in I}u_i^{\top}u_{i_{-}}.
    \label{eq:lrsa-anchors}
\end{equation}
The preference
$\delta_i=u_i^{\top}u_{i_{+}}-u_i^{\top}u_{i_{-}}$ measures how strongly
member $i$ favors anchor $i_{+}$ over anchor $i_{-}$. We stably sort
members by decreasing $\delta_i$ and assign the first
\begin{equation}
    n_{\mathrm{left}}(I)=C_{\mathrm{grp}}
    \left\lfloor\frac{\lceil |I|/C_{\mathrm{grp}}\rceil}{2}\right\rfloor
    \label{eq:lrsa-split}
\end{equation}
members to one child and the remainder to the other. Recursion stops when
$|I|\leq C_{\mathrm{grp}}$. With full
buckets and the stated divisibility conditions, every leaf group has
$C_{\mathrm{grp}}$ members.

After grouping, let $G$ denote the set of member positions in a group.
We compute its center from the original, unnormalized indexing keys:
\begin{equation}
    \mu_G=\frac{1}{|G|}\sum_{i\in G}k_i^{I}.
    \label{eq:lrsa-centers}
\end{equation}
Normalization guides membership without removing magnitude information
from the centers or subsequent candidate scoring. We pack
$N_{\mathrm{grp}}$ consecutive groups in recursive output order into page
$j$. Let $\mathcal{G}_j$ denote the collection of groups on page $j$.
Its member set is $\mathcal{P}_j=\bigcup_{G\in\mathcal{G}_j}G$. Each page thus holds
$C_{\mathrm{page}}$ members described by $N_{\mathrm{grp}}$ centers,
one per group. Retaining these
small-group centers avoids collapsing the whole page into a single mean,
so a relevant subgroup can influence retrieval even when other groups on
the page are less relevant.

Member keys are stored contiguously within each page, with centers stored
separately in the corresponding group order. A position map records the
original indexing position of the member in each page slot. This layout
combines fine-grained representations with regular page access while preserving the identities
needed for causal filtering and final attention lookup.

\subsection{Coarse-to-Fine Retrieval and Candidate Selection}
\label{sec:lrsa-retrieval}

Retrieval separates the choice of historical pages from the choice of
individual positions for attention. For query $t$, coarse selection retrieves
$K_{\mathrm{page}}$ pages from the published index. Fine selection combines
the members of these pages with sink, sliding-window, and unindexed
candidates, scores each candidate using the original scoring function,
and selects the $K$ highest-scoring positions for attention.

\emph{Coarse page selection.}
Let $\mathcal{J}_E$ be the set of page indices in the published prefix ending
at $E$. For each group $G$ on page $j$, we score its center $\mu_G$ using
the same query heads and weights as in Eq.~\eqref{eq:dsa-score}. The group
score $s^{\mathrm{grp}}_{t,G}$ and page score $s^{\mathrm{page}}_{t,j}$ are
\begin{equation}
\begin{aligned}
    s^{\mathrm{grp}}_{t,G}
    &=\sum_{h=1}^{H_I}w_{t,h}\,
      \operatorname{ReLU}\!\left((q_{t,h}^{I})^{\top}\mu_G\right),\\
    s^{\mathrm{page}}_{t,j}
    &=\max_{G\in\mathcal{G}_j}s^{\mathrm{grp}}_{t,G}.
\end{aligned}
\label{eq:lrsa-coarse-scores}
\end{equation}
Here $\mathcal{G}_j$ is the collection of groups on page $j$, defined in
Section~\ref{sec:lrsa-build}. Taking the maximum allows a high-scoring group
to bring its page into refinement without averaging across groups. This
avoids dilution by lower-scoring groups that could otherwise cause the
page to be overlooked during coarse selection. We select pages globally
across all published buckets:
\begin{equation}
    \mathcal{J}_t=\operatorname{Top}_{\min(K_{\mathrm{page}},|\mathcal{J}_E|)}
    \{s^{\mathrm{page}}_{t,j}:j\in\mathcal{J}_E\}.
    \label{eq:lrsa-page-selection}
\end{equation}
The selected page indices form $\mathcal{J}_t$. Center-based scores guide
page selection, while the subsequent refinement stage uses individual key
scores to determine the final attention positions.

\emph{Fine candidate selection.}
The selected pages provide historical candidates, which we combine with
sink positions, the sliding window, and visible positions beyond the
published index. Let $n_t$ be the number of positions visible to query $t$,
$S$ the sink length, and $W$ the window length. The three directly included
sets are
\begin{equation}
\begin{aligned}
    \mathcal{S}_t &= \{1,\ldots,\min(S,n_t)\},
        &&\text{(sink)},\\
    \mathcal{W}_t &= \{\max(1,n_t-W+1),\ldots,n_t\},
        &&\text{(sliding window)},\\
    \mathcal{U}_t &= \{i:E<i\leq n_t\},
        &&\text{(unindexed suffix)}.
\end{aligned}
\label{eq:lrsa-fixed-candidates}
\end{equation}
Their union $\mathcal{F}_t=\mathcal{S}_t\cup\mathcal{W}_t\cup\mathcal{U}_t$
contains each position only once. Using $\mathcal{P}_j$ for the member
positions of page $j$, the complete candidate set is
\begin{equation}
    \mathcal{C}_t=\mathcal{F}_t\cup
    \left\{i\in\bigcup_{j\in\mathcal{J}_t}\mathcal{P}_j:
    1\leq i\leq\min(E,n_t)\right\}.
    \label{eq:lrsa-candidate-union}
\end{equation}
The union removes duplicate positions shared by the retrieved pages and
the directly included ranges. We identify these overlaps using the stored
original positions, since directional grouping changes the temporal order
of members within each page.

Each candidate is then scored using its actual indexing key $k_i^{I}$ and
the original scoring function $s_t(i)$ from Eq.~\eqref{eq:dsa-score}. The final selected
positions are
\begin{equation}
    \mathcal{A}_t=\operatorname{Top}_{\min(K,|\mathcal{C}_t|)}
    \{s_t(i):i\in\mathcal{C}_t\}.
    \label{eq:lrsa-refinement}
\end{equation}
All candidates compete in this single Top-$K$. Sink, window, and unindexed
positions are always included in the candidate set, but none receives a
reserved place in $\mathcal{A}_t$. The candidate count satisfies
$|\mathcal{C}_t|\leq C_{\mathrm{page}}K_{\mathrm{page}}+|\mathcal{F}_t|$.
We use $S=64$ sink positions and a sliding window of $W=512$
indexing positions by default. The page budget $K_{\mathrm{page}}$ controls how many historical
pages enter refinement, while $K$ controls how many positions are selected
from the combined candidate set for attention.

During prefill, candidates shared by overlapping directly included ranges
are stored once, with per-query references and masks preserving each query's
visible set. Selected candidate slots are mapped to original logical
positions and expanded to token positions when required. The main
attention then computes its own scores and normalization over the
corresponding KV entries.

\section{Kernel Design and Execution Optimizations}
\label{sec:gpu}

The LRSA implementation organizes indexing keys and group centers
in a paged layout, with a member-position map linking reordered keys to
their original logical positions. This layout supports recursive index
construction and page-based retrieval, with kernel fusion reducing their
execution overhead. During refinement, queries within a prefill chunk share
storage for overlapping fixed candidates, reducing repeated data transfers.
This section describes how these storage and execution choices implement
the retrieval procedure.

\subsection{Paged Index Storage and Logical Position Mapping}
\label{sec:gpu-layout}

The grouping procedure in Section~\ref{sec:lrsa-build} defines which keys
belong to each retrieval page. We materialize these pages in a reordered
member pool, where each reordered physical page contains $C_{\mathrm{page}}$ keys
organized as $N_{\mathrm{grp}}$ complete groups. Construction copies the
source FP8 keys and their scales
into this pool according to the final grouping permutation. Keys within
a reordered physical page occupy consecutive slots, but different pages
may be allocated at noncontiguous locations. Refinement reads the indexing
keys from the reordered pages selected by coarse retrieval.

Within each reordered page, the scoring backend stores FP8 vector elements
and one FP32 scale per vector. The vector elements form a contiguous block
followed by the scales.
With $C_{\mathrm{page}}=64$ and indexing dimension $d_I=128$, the packed
layout is
\begin{equation}
    \underbrace{64\times128\ \text{bytes}}_{\text{FP8 keys: }8192\ \text{bytes}}
    \;\Vert\;
    \underbrace{64\times4\ \text{bytes}}_{\text{FP32 scales: }256\ \text{bytes}},
    \label{eq:gpu-page-layout-example}
\end{equation}
where $\Vert$ denotes concatenation in memory. Each page occupies
$D_{\mathrm{page}}=8448$ bytes.

Centers use the same packed format. Each member page contains
$N_{\mathrm{grp}}$ groups, each represented by one center. These
$N_{\mathrm{grp}}$ centers are stored consecutively so that their scores
can be reduced to a single member-page score. A \emph{center storage page}
holds $C_{\mathrm{page}}$ centers and therefore describes multiple member
pages. Coarse selection reads the centers, while refinement reads the keys
in the selected reordered member pages. With the default capacities, one
bucket produces 16 member pages and two center storage pages.

To preserve each key's original logical indexing position after
reordering, we construct a member-position map alongside the reordered
pages. When a key is copied to reordered physical
page $p_{\mathrm{reord}}$ at within-page slot $o$, the map records its
original logical position $i$:
\begin{equation}
    i=\pi(p_{\mathrm{reord}},o).
    \label{eq:gpu-position-map}
\end{equation}
This map recovers each historical candidate's logical position from its
reordered page and slot. Before final Top-$K$, these positions identify
retrieved members that also appear in the sink or sliding-window ranges. Masking
these historical duplicates retains one candidate per logical position.
After Top-$K$ selection, the same map converts the selected historical
slots into logical positions. The attention backend uses these positions to access the corresponding
KV entries and compute attention.

\subsection{Recursive Index Construction and Kernel Fusion}
\label{sec:gpu-construction}

Index construction consists of initialization, recursive partitioning,
and final assembly. Initialization reads a complete bucket of $B$ FP8 indexing
keys, dequantizes them using their associated scales, and normalizes the
vectors for directional grouping. Recursive partitioning then divides the
bucket into groups of $C_{\mathrm{grp}}$ members.
Assembly copies the keys into the resulting page order, records their
logical positions, and computes the group centers.

A straightforward implementation of one partitioning round uses six
successive kernel launches. Within each current group, it (1) computes dot
products with the reference member, (2) selects the first anchor $i_-$,
(3) computes dot products with that anchor, (4) selects the second anchor
$i_+$, (5) computes each member's directional preference $\delta_i$, and
(6) stably sorts the members by preference. The sorted members are split into two
capacity-balanced child groups for the next round.
All groups at the same recursion depth are processed in parallel; the next
round begins once the current round completes.

For $B=1024$ and $C_{\mathrm{grp}}=8$, the group size decreases as
\begin{equation}
    1024\rightarrow512\rightarrow256\rightarrow128
    \rightarrow64\rightarrow32\rightarrow16\rightarrow8.
    \label{eq:gpu-partition-rounds}
\end{equation}
These seven rounds produce 128 leaf groups of eight members each. With
one launch per stage across all groups at a given depth, the straightforward
schedule requires 42 partitioning launches per bucket, excluding
initialization and assembly. We reduce data movement by updating only
member indices during recursion, and reduce kernel launches by fusing
stages according to group size.

We keep normalized vectors in their original bucket order and update only
integer member indices between rounds. Let $\Pi_{\mathrm{in}}$ map the
current list positions to original bucket members. For a position $z$
in the output list, let $\rho(z)$ denote the input position of the member placed
at $z$ by the within-group sort. Each round writes
\begin{equation}
    \Pi_{\mathrm{out}}[z]=\Pi_{\mathrm{in}}[\rho(z)].
    \label{eq:gpu-permutation-update}
\end{equation}
Sorting reads member indices from $\Pi_{\mathrm{in}}$ and writes the
updated order to $\Pi_{\mathrm{out}}$. Separate input and output buffers
prevent newly written indices from overwriting those that other threads
still need to read. The buffers exchange roles for the next round. Updating
only member indices avoids moving full key vectors at every recursion depth.

We choose the fusion granularity according to group size. Fusing an entire
partitioning round for a large group would increase the working set held
by each cooperative thread array (CTA), raising register pressure and
potentially causing register spilling that degrades performance. For large
groups, we therefore fuse only two pairs of stages: first-anchor selection
with the subsequent anchor scoring, and second-anchor selection with
preference scoring, resulting in four kernel launches per round. For small
groups, the smaller working set allows a single CTA to fuse all six stages
into one kernel, reusing loaded vectors and avoiding intermediate score
writes between stages.
With $B=1024$ and $C_{\mathrm{grp}}=8$, group sizes 1024, 512, 256, and
128 use the four-launch schedule, while sizes 64, 32, and 16 use the
fully fused schedule. The seven partitioning rounds therefore require
19 launches. This group-size-dependent design balances register demand
and parallelism against launch overhead and intermediate memory traffic.

Once partitioning reaches the leaf groups, a single assembly pass follows
the final permutation to write reordered member pages and their position maps.
The same loaded keys are used to compute group centers from their
dequantized, unnormalized values. Member keys retain their source FP8 values
and scales, while centers are quantized with their own scales. The completed
bucket is then published through the asynchronous schedule described in
Section~\ref{sec:lrsa-overview}.

\emph{Fused Page Reduction and Selection.}\label{sec:gpu-routing} After center scoring, coarse retrieval computes page scores, selects
pages, and writes their reordered physical page indices for refinement.
We fuse these three operations into a single routing kernel, assigning
one CTA to each query. For each member page, the CTA takes the maximum
of its $N_{\mathrm{grp}}$ center scores to obtain the page score. It then
selects the Top-$K_{\mathrm{page}}$ pages using block-level radix sorting
and directly writes their reordered physical page indices for refinement.
This fusion of max reduction, Top-$K_{\mathrm{page}}$ selection, and output
writing avoids intermediate page-score writes to global memory and reduces
kernel launches.

\subsection{Shared Candidate Storage for Chunked Prefill}
\label{sec:gpu-sharing}

Refinement jointly scores retrieved historical keys and the fixed
candidates comprising sink, sliding-window, and unindexed keys. These two
sets initially reside in different layouts: historical keys are stored
in the reordered member pool, while fixed candidates remain in the source
indexing-key storage. The DeepGEMM paged scoring backend accepts a single
key-pool base address, so we copy the fixed candidates into a reserved
region of the reordered pool. Both sets can then be scored through the
same paged interface. During chunked prefill, however, neighboring
queries use largely overlapping sink and fixed-tail ranges. Preparing a
separate copy for each query would repeatedly transfer the same keys and scales.

To avoid these repeated transfers, we construct a single shared copy of
the fixed candidates for one prefill chunk. Consider its $R$ consecutive
queries with nondecreasing visible lengths $n_1,\ldots,n_R$ and a common
published index boundary $E$. Each query's fixed candidates consist of a
sink prefix and a fixed tail covering its sliding window and visible
unindexed suffix. After excluding overlap with the sink prefix, let
$\tau_1$ denote the logical boundary immediately before the first query's
fixed tail. This tail contains positions $\{\tau_1+1,\ldots,n_1\}$.
Across the chunk, the first query determines the start of the required
fixed-tail range, while the last query determines the
final visible position $n_R$. The logical positions to copy are therefore
\begin{equation}
    \mathcal{F}_{\mathrm{shared}}
      =\underbrace{\{1,\ldots,\min(S,n_R)\}}_{\text{sink positions}}
       \;\cup\;
       \underbrace{\{\tau_1+1,\ldots,n_R\}}_{\text{fixed-tail positions}}.
    \label{eq:gpu-shared-positions}
\end{equation}
This union covers the fixed candidates needed by every query in the chunk,
including the full visible unindexed suffix, and counts overlapping
positions only once.

We copy the keys and scales for $\mathcal{F}_{\mathrm{shared}}$
into a reserved region of the reordered pool, where they are stored as
shared pages in the same packed format as the historical member pages.
The shared region stores the sink portion followed by the remaining
fixed-tail portion. Copies are page-aligned, and each required source
page is transferred only once. Page alignment may include positions
outside the required ranges; query-specific boundaries exclude those
positions during refinement.

During refinement, each query references its selected reordered historical
pages and the shared pages covering its own sink and fixed-tail ranges.
These references allow queries to reuse the shared storage without creating
additional per-query copies. Query-specific lengths and masks exclude page
padding and future positions, while the logical
position map identifies historical members duplicated by the sink or
sliding-window ranges. Thus, the queries share stored keys but retain their
own candidate sets and scores.

The reduction in copying depends on the overlap among fixed candidate
ranges. Let $\mathcal{V}_t$ denote the source-page set covering the fixed candidates
of query $t$. With $D_{\mathrm{page}}$ bytes per packed page, the copied
data volumes for separate per-query copies and shared storage are
\begin{equation}
    D_{\mathrm{private}}
      =D_{\mathrm{page}}\sum_{t=1}^{R}|\mathcal{V}_t|,
    \qquad
    D_{\mathrm{shared}}
      =D_{\mathrm{page}}\left|\bigcup_{t=1}^{R}\mathcal{V}_t\right|.
    \label{eq:gpu-shared-copy}
\end{equation}
A source page needed by several queries contributes once to the shared
copy, instead of once per query.

\section{Runtime Index Management and Reuse in SGLang}
\label{sec:runtime}

The LRSA runtime in SGLang maintains index state
for each request and attention layer, coordinating asynchronous construction
with queries over the published index. Prefix matches in the original KV
pool enable reuse of complete bucket indices or assembled member pages,
while retention policies determine how long these resources remain
available to subsequent requests. This section describes how request
management, index reuse, and retention support ongoing inference while
balancing index storage against repeated construction.

\subsection{Request State and Asynchronous Index Construction}
\label{sec:runtime-integration}

The attention backend connects retrieval to model execution. Once the
current layer's indexing keys and original KV entries are available,
LRSA retrieves from the published index and returns logical positions
to the model's attention backend. After attention, the backend submits
construction tasks for newly filled buckets. Model projections and the
original KV cache retain their existing roles; the additional runtime
state tracks the index used to select attention positions.

Prefix matching in the original KV cache identifies reusable source state,
while a lookup in the derived index determines whether an existing bucket
record can be reused without reconstruction. Each query's visible length determines
which positions may participate in retrieval. The original KV cache, reordered
index pool, and request-local page references have distinct
roles and lifetimes.

For a fixed request, layer, and indexing-key head, we represent the
retrieval state as
\begin{equation}
    \mathcal{R}=\left(\mathrm{id},\mathcal{L},
        \mathsf{T}_{\mathrm{mem}},\mathsf{T}_{\mathrm{ctr}},
        E,\mathcal{E}\right),
    \label{eq:runtime-state}
\end{equation}
where $\mathrm{id}$ uniquely identifies the request,
$\mathcal{L}$ is the ordered list of bucket records,
$\mathsf{T}_{\mathrm{mem}}$ and $\mathsf{T}_{\mathrm{ctr}}$ contain
member-page and center-page references, and $\mathcal{E}$ contains the
completion events associated with index updates. As in
Section~\ref{sec:lrsa-overview}, $E$ marks the end of the published index
prefix available to queries. Each bucket record associates
its logical range with its centers, reordered member pages, and position
maps. Queries access these records and the published boundary together
as a consistent index snapshot.
Visible positions beyond $E$ remain fixed refinement candidates, as defined in
Section~\ref{sec:lrsa-retrieval}.

Index updates begin on the CPU, which checks tensor shapes and available
capacity, allocates new buckets, and prepares source-page metadata.
Host-side locks protect record allocation and updates to request containers
and record metadata. After the current
layer's attention, the CPU submits the prepared work to a dedicated CUDA
stream. CUDA event dependencies ensure that preceding operations on shared
construction workspaces and the reordered-page allocator, including its
free-page stack, complete before this stream accesses those resources.
The GPU then groups the keys, assembles pages, and verifies potential
matches in the reordered pool.
It atomically increments reference counts for reusable pages and allocates storage
from the free pool for unmatched pages. Allocation, publication, and
reclamation follow the same cross-stream ordering so that shared resources
remain valid throughout their use.

Once the new index records have been published, the construction stream
records a completion event. While construction proceeds, the foreground stream
continues with subsequent layers and independent computation, enabling
the asynchronous overlap described in Section~\ref{sec:lrsa-overview}.
Before the next query batch uses the updated index at the same layer,
its stream waits on that event and then reads the page references, centers,
member position maps, and $E$ from the completed update.

\subsection{Prefix-Based Reuse of Bucket Indices and Member Pages}
\label{sec:runtime-reuse}

Following a prefix match in the original KV pool, index reuse can occur
before bucket construction or after page assembly. These two forms of
reuse avoid different costs: reusing a complete bucket avoids rebuilding
its index, while sharing assembled member pages reduces duplicate storage.

\emph{Complete bucket reuse.} The matched original KV prefix must extend
through the end of the bucket, covering both the entire bucket and its
preceding context. We associate each such cached prefix with its complete
bucket record. For bucket $\mathcal{B}_m$, the reuse identity is
\begin{equation}
    \mathcal{I}_m=\left(\xi_m,\mathcal{B}_m,\lambda,h_I,
        \theta\right),
    \label{eq:runtime-bucket-identity}
\end{equation}
where $\xi_m$ identifies the logical prefix through the end of the bucket,
$\lambda$ and $h_I$ identify the layer and indexing-key head, and $\theta$
specifies the model state, key representation, and construction configuration.
When the prefix match covers the complete bucket and the remaining
compatibility fields agree, the backend checks that the associated record
and its index resources remain valid. It then acquires a reference,
establishes the record's completion dependency, and attaches the existing
center-page and member-page references to the request. This path skips grouping, center construction,
and page assembly. If the associated index has been reclaimed, the bucket
must instead be rebuilt from valid source keys. Complete bucket reuse
therefore requires both a matching original
KV prefix and a valid index record for that bucket.

\emph{Page reuse after construction.} When a bucket within the matched
prefix requires construction, assembly produces temporary member pages
and their position maps. We generate fingerprints for each page from its
packed keys, scales, and member positions, then compare them with the
fingerprints of live pages to identify potential matches. A full comparison
of the page contents and position maps confirms equality.
For each confirmed match, the bucket acquires a reference to the existing
page; unmatched pages receive new storage. Each bucket retains its own center records and references
the resulting member pages. This path still performs grouping, assembly,
and content checking, but avoids storing another copy of each matching
member page.

\subsection{Index Retention Policies and Storage Tradeoffs}
\label{sec:runtime-retention}

The reuse mechanisms determine which index resources can be shared,
while the retention policy determines how long they remain available.
We design two policies that differ in whether the prefix cache of the
original KV pool retains bucket records and their associated reordered
index resources after active requests finish or allows them to be
reclaimed. Both policies support page-level sharing and, with a valid
prefix-to-index association, complete bucket reuse.

\emph{Request-scoped retention.} Active requests hold bucket records until
their GPU operations complete. Once the last active holder releases a
record, the record and any index resources no longer held elsewhere can be reclaimed,
even if the matching prefix remains cached in the original KV pool.
This keeps the live index working set tied to active requests and frees
capacity for new buckets rather than retaining indices for inactive
prefixes. A new request that matches a cached prefix in the original KV
pool can reuse its compatible bucket record while that record remains
held. Once the record has been reclaimed, however, the same prefix match
requires index reconstruction. Repeated requests can therefore incur
repeated grouping, center construction, and page matching.

\emph{Prefix-cache retention.} The prefix cache of the original KV pool
holds the bucket records associated with compatible cached prefixes.
After a request completes, the cache retains the centers, reordered member
pages, and position maps for subsequent requests. This extends complete
bucket reuse beyond the active request lifetime, allowing a valid hit to
skip construction while both the original KV prefix and its index remain cached.
Evicting that prefix from the original KV pool releases the cache's
references to its bucket records. Reclamation waits until all remaining
holders release their references and outstanding GPU accesses complete.
This longer reuse window comes at the cost of retaining index resources
for inactive prefixes. Under a fixed GPU memory budget, this storage
competes with original KV entries,
active indices, and execution workspaces.

The choice therefore depends on prefix reuse in the original KV pool and
available capacity. Request-scoped retention limits inactive index residency
but may require repeated construction; prefix-cache retention preserves
more reusable work but can reduce the capacity available to other requests.
A separate index-cache
budget can limit index residency while preserving cached prefixes in the
original KV pool, with missing indices rebuilt on subsequent prefix hits.

\section{Experiments and Results}
\label{sec:experiments}

We evaluate LRSA's downstream task performance and inference efficiency.
We further analyze how index organization affects retrieval quality and
how kernel optimizations and asynchronous index maintenance contribute
to execution efficiency.

\subsection{Setup}
\label{sec:experimental-setup}

\noindent\emph{Models, datasets, and baselines.}
We evaluate task quality with Qwen3-4B-Instruct-2507
\citep{yang2025qwen3} and Hunyuan-7B-Instruct-0124 on representative datasets: NarrativeQA \citep{kocisky2017narrativeqa}, MuSiQue \citep{Trivedi2021MM}, HotpotQA \citep{yang2018hotpotqa},
TriviaQA \citep{Joshi2017TriviaQAAL}, TREC \citep{Li2002LearningQC}, 2WikiMQA \citep{Ho2020ConstructingAM}, DuReader \citep{He2017DuReaderAC}, LSHT, QMSum \citep{Zhong2021QMSumAN}, and Qasper \citep{Dasigi2021ADO}.
These tasks cover document question answering, multi-hop reasoning,
few-shot classification, and summarization. We report each task's official
score on a scale from zero to one hundred, together with the unweighted
mean across tasks. We compare against dense attention, which attends to
all causally visible positions, and DSA \citep{deepseek2025v32}, which
exhaustively scores all causally visible keys and selects the exact
Top-$K$ positions. Under the same indexing function and selection budget,
DSA maximizes the sum of selected-key scores, providing an optimal
reference for the key-selection objective. We evaluate how closely LRSA
approaches this reference while reducing retrieval cost. DSA and LRSA use
a matching final selection budget of 2048 tokens for task-quality
evaluation and budgets of 2048, 1024, and 512 tokens for the end-to-end
comparisons.

\noindent\emph{Implementation details.}
We run our experiments on NVIDIA H20 GPUs. The end-to-end implementation
runs in SGLang, stores model weights and original KV entries in BF16,
and gathers selected KV entries for attention computation with
FlashAttention-3.
By default, the bucket capacity is $B=1024$. Within each bucket,
capacity-balanced grouping based
on cosine similarity places $C_{\mathrm{grp}}=8$ keys in each group.
Each page contains $N_{\mathrm{grp}}=8$ complete groups, giving a page
capacity of $C_{\mathrm{page}}=64$ members. A group center is the mean of
its members' unnormalized indexing keys; both keys and centers are
stored in FP8 with FP32 scales. Coarse retrieval scores each page by the
maximum score of its group centers. We set the reference page budget $K_{\mathrm{page}}$ to the number of
pages needed to achieve a mean all-position attention-mass recall (AMR)
of 95\% across all queries. AMR is defined in
Section~\ref{sec:retrieval-ablations}. Refinement applies the original
scoring function to the retrieved page members and candidates from
$S=64$ sink positions, a sliding window of $W=512$ indexing positions,
and the visible unindexed suffix, then selects the final Top-$K$ positions.
These settings remain fixed unless explicitly varied in the
corresponding experiment.

\subsection{Performance on Downstream Tasks}
\label{sec:task-quality}

\begin{table}[t]
\centering
\small
\setlength{\tabcolsep}{5pt}
\caption{Performance of dense attention, DSA, and LRSA across ten downstream
tasks. DSA and LRSA use the same final selection budget of 2048 tokens.}
\label{tab:task-quality}
\begin{tabular}{lrrrrrr}
\toprule
\multirow{2}{*}{Task} & \multicolumn{3}{c}{Qwen3-4B-Instruct-2507} & \multicolumn{3}{c}{Hunyuan-7B-Instruct-0124} \\
\cmidrule(lr){2-4}\cmidrule(lr){5-7}
 & Dense & DSA & LRSA & Dense & DSA & LRSA \\
\midrule
NarrativeQA & 27.45 & 27.94 & 27.26 & 22.98 & 22.64 & 23.17 \\
MuSiQue & 25.40 & 22.72 & 23.23 & 29.91 & 28.95 & 28.72 \\
HotpotQA & 58.95 & 57.06 & 57.54 & 51.38 & 53.31 & 53.38 \\
TriviaQA & 85.23 & 84.47 & 84.27 & 79.87 & 79.78 & 79.42 \\
TREC & 74.00 & 74.50 & 74.50 & 63.50 & 62.50 & 62.00 \\
2WikiMQA & 43.53 & 44.42 & 44.30 & 45.59 & 46.61 & 46.20 \\
DuReader & 26.64 & 26.54 & 26.58 & 31.27 & 31.27 & 31.50 \\
LSHT & 43.50 & 47.00 & 46.00 & 27.16 & 26.37 & 26.25 \\
QMSum & 22.37 & 22.75 & 22.52 & 23.67 & 24.39 & 24.26 \\
Qasper & 44.86 & 43.72 & 44.18 & 44.29 & 43.45 & 43.59 \\
\midrule
Average & 45.19 & 45.11 & 45.04 & 41.96 & 41.93 & 41.85 \\
\bottomrule
\end{tabular}
\end{table}

We evaluate whether LRSA preserves downstream task performance when
approximating DSA's optimal selection under the indexing objective.
DSA exhaustively scores all causally visible keys and returns the exact
Top-K positions at the specified budget. LRSA narrows
the historical candidate set through page selection, then applies the same
scoring function to the retrieved members and fixed candidates during
refinement. Comparing the two at the same final budget therefore measures
how closely LRSA's approximate retrieval preserves the downstream task
performance obtained with exhaustive selection.

Table~\ref{tab:task-quality} shows that LRSA closely matches DSA's average
performance on both models at a final selection budget of 2048 tokens.
The absolute difference in the ten-task average is at most 0.08 points
on a hundred-point scale. On Qwen3-4B-Instruct-2507, LRSA scores 45.04
compared with DSA's 45.11, a difference of 0.07 points. On
Hunyuan-7B-Instruct-0124, LRSA scores 41.85 compared with DSA's 41.93,
a difference of 0.08 points. These results indicate that LRSA's page-based
retrieval closely preserves the average downstream performance obtained
with exhaustive key selection.

\subsection{Impact of Kernel Fusion and Candidate Sharing}
\label{sec:kernel-experiments}

We examine how kernel fusion and candidate sharing reduce the cost of
index construction and candidate preparation. We measure partitioning
latency under different fusion schedules and copying latency with and
without shared fixed-candidate storage, evaluating the optimizations in
Sections~\ref{sec:gpu-construction} and~\ref{sec:gpu-sharing}.

\begin{figure}[!htbp]
\centering
    \includegraphics[width=0.49\linewidth]{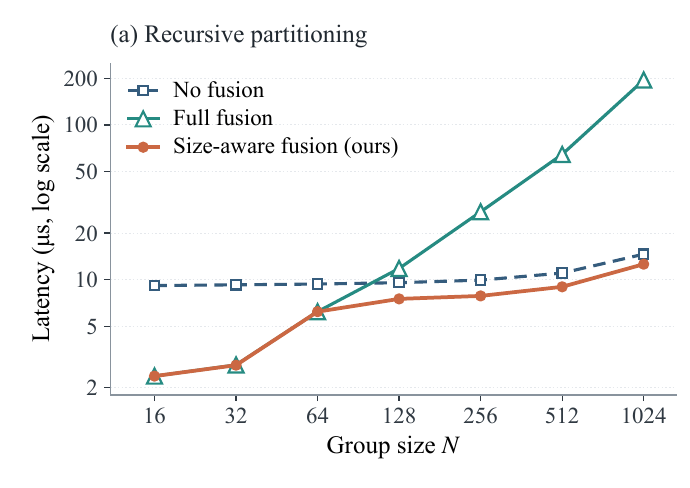}\hfill
    \includegraphics[width=0.49\linewidth]{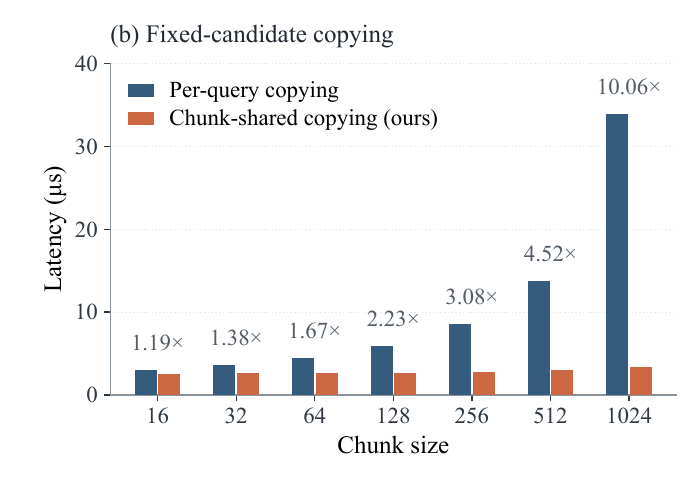}
    \caption{Latency of recursive partitioning and fixed-candidate copying.
    (a) Partitioning latency for three fusion schedules as group size
    varies, with the total number of members per round held fixed.
    (b) Candidate-copying latency as chunk size varies, comparing
    per-query copies with a shared copy. Labels in (b) show the speedup
    from sharing. Panel (a) uses a logarithmic latency axis; panel (b)
    uses a linear axis starting at zero.}
    \label{fig:kernel-optimizations}
\end{figure}

\noindent\emph{Group-size-dependent fusion.}
We compare three schedules for index partitioning: no fusion, full
fusion, and our size-aware fusion. Our schedule selectively fuses stages
for large groups and fuses the entire round for small groups. We vary
the group size $N$ while holding the total number of members processed
per round fixed, allowing us to compare the schedules at the same workload.

Figure~\ref{fig:kernel-optimizations}(a) shows that the benefit of fusion
depends on group size. At $N=1024$, full fusion takes
195.06 microseconds, compared with 14.61 microseconds without fusion
and 12.59 microseconds with our size-aware schedule. At $N=128$, the
corresponding latencies are 11.81, 9.56, and 7.51 microseconds.
For smaller groups ($N\leq64$), full fusion becomes faster than unfused
execution. Our schedule uses full fusion for these groups, and its
latency closely matches the fully fused baseline. At $N=16$, it reduces
latency from 9.16 microseconds without fusion to
2.38 microseconds. Across all measured group sizes, our schedule achieves a speedup of $1.16\times$ to $3.85\times$
over no fusion. These results support fusing the entire round for small
groups while using selective fusion for larger groups.

\noindent\emph{Shared fixed-candidate copying.}
We compare copying fixed candidates separately for each query with
preparing their union once for the entire chunk. We vary chunk size
from 16 to 1024 queries. Figure~\ref{fig:kernel-optimizations}(b) shows that per-query
copying latency rises from 2.963 to 33.916 microseconds, whereas shared
copying latency increases only from 2.492 to 3.370 microseconds.
The resulting speedup increases from $1.19\times$ to $10.06\times$,
reaching $3.08\times$ at 256 queries and $4.52\times$ at 512 queries.
This growing advantage comes from reusing keys in the overlapping sink
and tail ranges of neighboring queries. A shared copy serves more queries
as the chunk grows, avoiding repeated transfers and limiting the increase
in copying latency.

\subsection{Efficiency of Long-Context Inference}
\label{sec:attention-efficiency}
\label{sec:full-model-prefill}

We evaluate how LRSA's retrieval design affects the cost of selecting
keys, computing attention, and executing the full model across context
lengths from 64K to 1024K tokens. For key selection, we report LRSA's
speedup over DSA. For attention computation, we include both key selection
and sparse attention and compare DSA and LRSA with full attention.
For each comparison, speedup is the ratio of the reference method's
execution time to that of the evaluated method.

\begin{figure}[!htbp]
\centering
    \includegraphics[width=\linewidth]{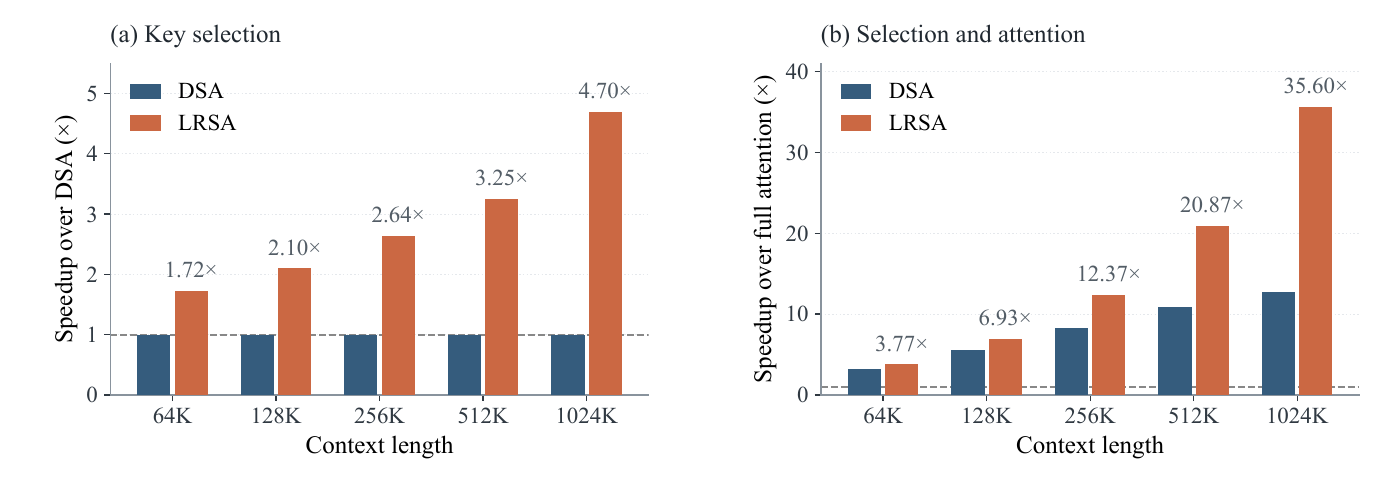}
    \caption{Efficiency of key selection and attention across context lengths.
    (a) Key-selection speedup of LRSA over DSA.
    (b) Speedups of DSA and LRSA over full attention; both sparse methods
    include key selection and attention computation. Labels above LRSA
    bars give its speedup over the reference in each panel. Dashed lines
    mark equal execution time ($1\times$).}
    \label{fig:attention-efficiency}
\end{figure}

\begin{wraptable}{r}{0.46\textwidth}
\centering
\caption{End-to-end speedup of LRSA over DSA across context lengths and
final selection budgets.}
\label{tab:full-model-timing}
\begin{tabular}{cccc}
\toprule
\multirow{2}{*}{\shortstack[c]{Context\\length}} & \multicolumn{3}{c}{Selection budget} \\
\cmidrule(lr){2-4}
 & 2048 & 1024 & 512 \\
\midrule
64K & $1.144\times$ & $1.130\times$ & $1.125\times$\\
128K & $1.249\times$ & $1.333\times$ & $1.329\times$\\
256K & $1.667\times$ & $1.702\times$ & $1.738\times$\\
512K & $1.638\times$ & $1.762\times$ & $1.755\times$\\
1024K & $1.865\times$ & $1.957\times$ & $1.984\times$\\
\bottomrule
\end{tabular}
\end{wraptable}

Figure~\ref{fig:attention-efficiency}(a) shows that LRSA achieves
key-selection speedups over DSA of $1.72\times$ at 64K and $4.70\times$
at 1024K.
DSA scores the full visible context, whereas LRSA uses page-level
retrieval to restrict fine-grained scoring to selected historical members
and fixed candidates. This smaller search space becomes increasingly
beneficial as the context grows. When both selection and sparse attention
are included, LRSA achieves speedups over full attention of
$3.77\times$ at 64K and $35.60\times$ at 1024K; the corresponding DSA
speedups are $3.25\times$ and $12.70\times$
(Figure~\ref{fig:attention-efficiency}(b)). These results show that
LRSA's selection gains remain evident when the subsequent sparse
attention computation is included.

To examine how these gains translate to full-model inference, we also
compare DSA and LRSA over the same range of context lengths, using
matching final selection budgets of 2048, 1024, and 512 tokens.
The end-to-end measurement includes model execution together with index
construction and incremental maintenance, candidate preparation, position
handling, scheduling gaps, and synchronization. The resulting speedup
therefore reflects the net benefit of LRSA within full-model execution,
with index-management costs included.

Table~\ref{tab:full-model-timing} shows that LRSA is faster than DSA
across all evaluated context lengths and selection budgets. At 64K,
the speedup ranges from $1.125\times$ to $1.144\times$ across the
three budgets. It reaches $1.25\times$ to $1.33\times$ at 128K and
$1.67\times$ to $1.74\times$ at 256K. At 512K and 1024K, the ranges
are $1.64\times$ to $1.76\times$ and $1.86\times$ to $1.98\times$,
respectively. LRSA thus
retains an end-to-end advantage over DSA after accounting for index
maintenance, with the largest gains observed at the longest evaluated context.

\subsection{Ablation Studies}
\label{sec:retrieval-ablations}

\begin{figure}[!t]
\centering
    \includegraphics[width=\linewidth]{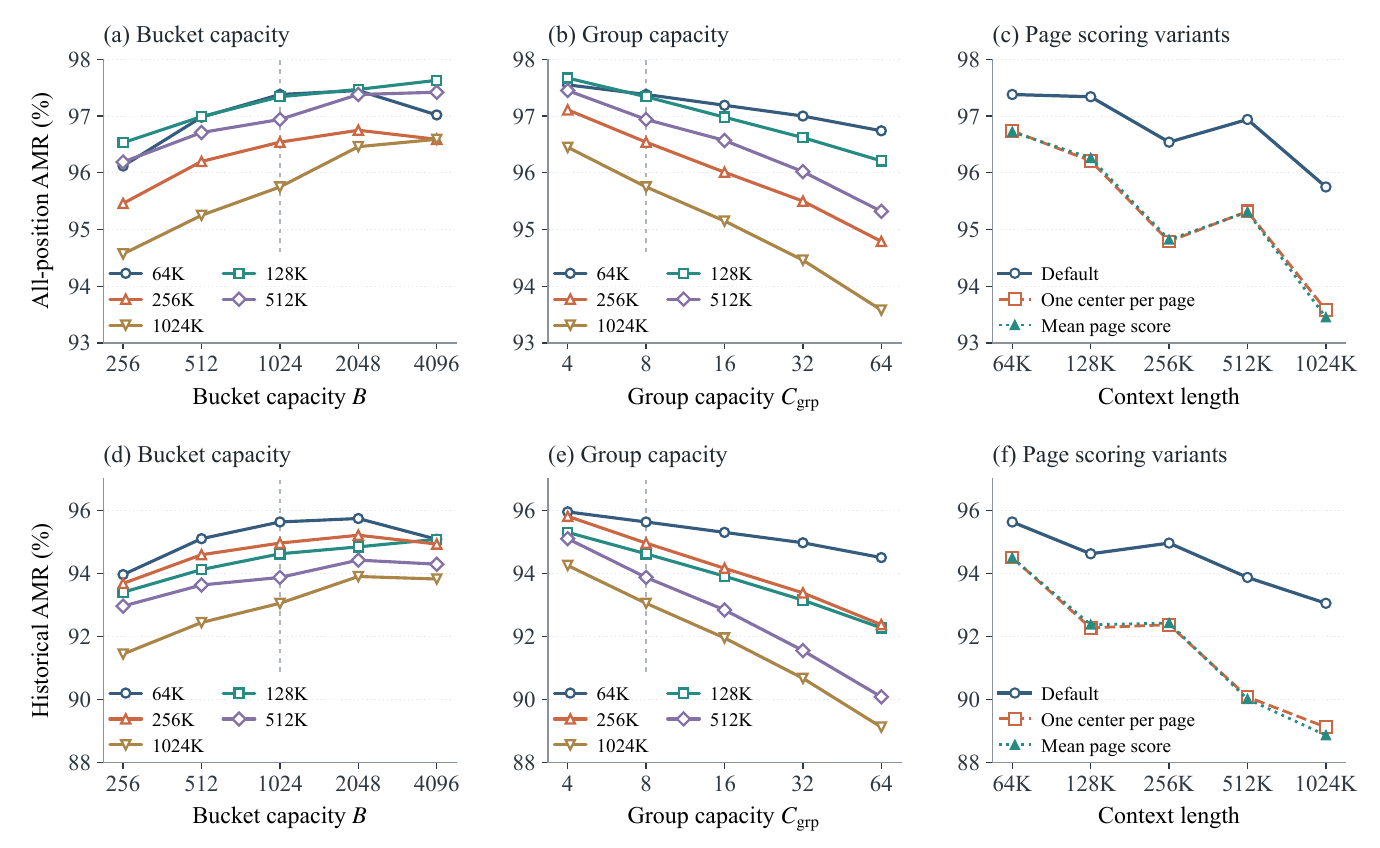}
    \caption{Ablations of index organization across five context lengths.
    The top row reports all-position AMR and the bottom row historical
    AMR. Panels (a, d) vary bucket capacity, and panels (b, e) vary group
    capacity. In these panels, colors denote context lengths and vertical
    dashed lines mark the default capacities. Panels (c, f) compare the
    default with two variants: one mean center per page and mean aggregation
    of group scores. Each variant changes only the corresponding design
    choice. Page capacity and selection budgets are held fixed within
    each comparison. Points report mean AMR.}
    \label{fig:ablation-quality}
\end{figure}

\begin{figure}[!t]
    \centering
    \includegraphics[width=\linewidth]{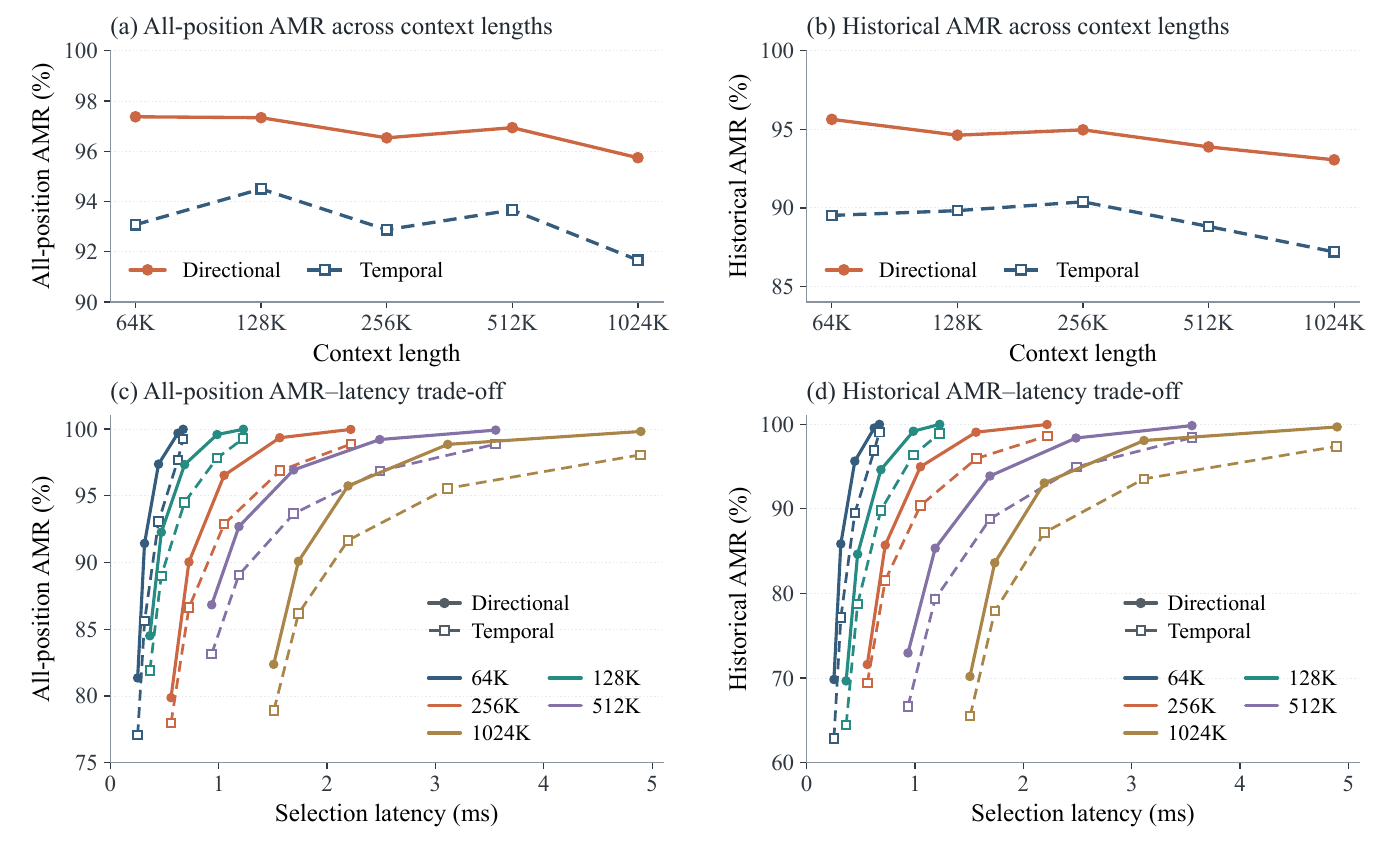}
    \caption{Ablation of directional grouping using temporal grouping as
    the control. The left and right columns report all-position and
    historical AMR, respectively. Panels (a, b) compare the two methods
    at matched reference page budgets, with colors distinguishing the
    methods. Panels (c, d) show the trade-off between AMR and selection
    latency as the page budget varies; colors denote context lengths,
    and each point corresponds to one page-budget setting. In all panels,
    solid lines with circles indicate directional grouping, and dashed
    lines with squares indicate temporal grouping.}
    \label{fig:grouping-comparison}
\end{figure}

We ablate the main choices in LRSA's index organization: bucket capacity,
group capacity, page representation, page-score aggregation, and directional
grouping. For each comparison in Figure~\ref{fig:ablation-quality}, we change one setting while holding
the queries, fixed candidates, page capacity, and selection budgets
constant. Figure~\ref{fig:grouping-comparison} further compares directional
and temporal grouping at matched page budgets and over a budget sweep.

\noindent\emph{Attention-mass recall.}
To make differences in retrieval quality more apparent, we use
attention-mass recall (AMR), which measures how much reference attention
mass is preserved by the selected positions. Weighting each recovered
position by its attention contribution makes AMR sensitive to the
importance of the keys retained by each index design.
Let $\mathcal D_q$ and $\mathcal V_q$ be the original token positions
selected by DSA and LRSA for query $q$, and let $a_{qhi}$ be the reference
attention weight at position $i$ for head $h$, obtained by normalizing
the attention logits over $\mathcal D_q$ with softmax. We define
\begin{equation}
    \operatorname{AMR}_{qh}
    =\sum_{i\in\mathcal D_q\cap\mathcal V_q}a_{qhi},
    \qquad
    \operatorname{AMR}^{\mathrm{hist}}_{qh}
    =\frac{\sum_{i\in\mathcal H_q\cap\mathcal V_q}a_{qhi}}
           {\sum_{i\in\mathcal H_q}a_{qhi}},
    \label{eq:experimental-amr}
\end{equation}
where $\mathcal H_q$ is the subset of $\mathcal D_q$ outside the sink
and sliding-window ranges. All-position AMR measures the fraction of DSA's
selected attention mass retained by LRSA. Historical AMR excludes the
contribution of the shared sink and sliding-window positions, making
differences among index designs more apparent without dilution by these
fixed candidates.

\noindent\emph{Bucket capacity.}
Figure~\ref{fig:ablation-quality}(a, d) shows that enlarging the bucket
initially improves recall, but the benefit does not increase monotonically
with capacity. Increasing $B$ from 1024 to 2048 improves historical AMR
at all five lengths by 0.11 to 0.85 percentage points. A further increase
to 4096 lowers historical AMR at 64K from 95.75\% to 95.09\%, while
raising it at 128K from 94.85\% to 95.09\%. Across lengths, $B=2048$
achieves a mean historical AMR of 94.83\%, compared with 94.65\% for
$B=4096$. These results suggest that a broader grouping scope can
improve retrieval, but the benefit depends on context length and does
not necessarily increase with bucket capacity.

\noindent\emph{Representative granularity.}
Figure~\ref{fig:ablation-quality}(b, e) shows that finer groups improve
recall while page capacity remains fixed. Reducing $C_{\mathrm{grp}}$
from 8 to 4 yields historical AMR gains of 0.32 to 1.23 percentage points across
the five lengths. Conversely, setting $C_{\mathrm{grp}}=64$, so that
one center represents the entire page, lowers historical AMR by 1.13
points at 64K and 3.93 points at 1024K relative to the default.
To isolate representative granularity from page membership, the
single-center variant in Figure~\ref{fig:ablation-quality}(c, f) keeps
the default member pages and replaces each page's group centers with
one mean center. This variant yields the same AMR as $C_{\mathrm{grp}}=64$
at all five lengths because further partitioning within a page changes
its group centers without changing its members. Both comparisons
support retaining fine-grained group centers to capture relevant subsets
of a page.

\noindent\emph{Page-score aggregation.}
Replacing max aggregation with mean aggregation retains the same groups
and centers but reduces historical AMR at every length
(Figure~\ref{fig:ablation-quality}(c, f)). The reduction grows from
1.14 percentage points at 64K to 4.19 points at 1024K. This variant differs
from the single-center variant: it averages the group scores rather than
their key vectors. Although the two produce similar AMR in these
experiments, both fall below the default. The results support ranking a
page by its strongest group response, so that a relevant group's score
is not diluted by less relevant groups on the same page.

\noindent\emph{Directional versus temporal grouping.}
We compare directional grouping with temporal grouping, which groups
consecutive keys in their original sequence order. The two methods use
the same bucket, group, and page capacities, page-scoring rule, fixed
candidates, and final selection budgets. We evaluate both on identical
queries and historical prefixes, first at matched page budgets and then
across a page-budget sweep.

At the reference page budgets, directional grouping improves
all-position AMR by 2.84 to 4.30 percentage points and historical AMR by
4.58 to 6.12 points across the five lengths
(Figure~\ref{fig:grouping-comparison}(a, b)). These are aggregate gains;
individual comparisons can still show a decrease in historical AMR.
The budget sweep complements this comparison by showing how retrieval
quality changes with selection latency
(Figure~\ref{fig:grouping-comparison}(c, d)). At 1024K, directional grouping with twice the reference page budget
achieves 98.078\% historical AMR in 3.1131 ms. Temporal grouping with
four times the budget reaches 97.383\% in 4.8908 ms. At these measured
operating points, directional grouping retains more attention mass
with a smaller page budget and a selection speedup of $1.57\times$.

\FloatBarrier

\subsection{Evaluating Overlap in Index Maintenance}
\label{sec:overlap-experiment}

We evaluate how much index-maintenance overhead can be hidden by
model execution between consecutive chunks. Keeping the model computation
fixed, we compare execution without index maintenance, with serial
maintenance, and with asynchronous maintenance at a single layer.
We also measure index construction in isolation to assess its cost
relative to the available execution window. We set bucket capacity equal
to chunk size and evaluate capacities of 256, 512, 1024, 2048, and 4096
across context lengths from 8K to 256K tokens.

\begin{figure}[!htbp]
    \centering
    \includegraphics[width=\linewidth]{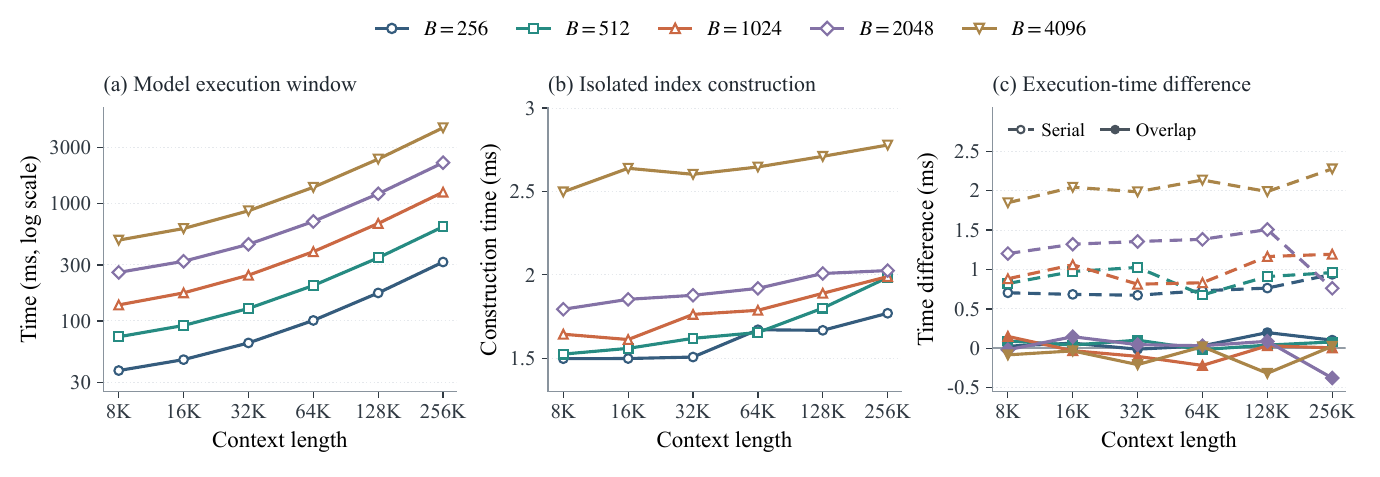}
    \caption{Overlap between index maintenance and model execution.
    Colors and marker shapes identify five bucket capacities.
    (a) Model-only execution window. (b) Isolated index-construction time.
    (c) Execution-time difference relative to the model-only baseline.
    In (c), dashed lines with open markers indicate serial maintenance,
    while solid lines with filled markers indicate overlap. All points
    show median measurements.
    The vertical axis is logarithmic in (a) and linear in (b, c).}
    \label{fig:maintenance-overlap}
\end{figure}

We time the interval from the completion of attention at the monitored
layer in one chunk to the start of attention at that layer in the next
chunk, including any wait for index publication. This gives
$T_{\mathrm{window}}$ for model-only execution,
$T_{\mathrm{serial}}$ with serial maintenance, and
$T_{\mathrm{overlap}}$ with asynchronous maintenance.
The isolated construction time $T_{\mathrm{build}}$ measures the same
index update without concurrent model computation. For each repeat $r$,
we compute the execution-time difference from the model-only baseline as
\begin{equation}
    \Delta T_m^{(r)}=T_m^{(r)}-T_{\mathrm{window}}^{(r)},
    \qquad m\in\{\mathrm{serial},\mathrm{overlap}\}.
    \label{eq:overlap-exposed-overhead}
\end{equation}
Differences are computed for each repeat before taking the median.
Context lengths refer to the chunk that consumes the updated index.

Figure~\ref{fig:maintenance-overlap}(a, b) shows that the measured model
execution windows substantially exceed isolated construction times across all
five bucket capacities. Median window durations range from 37.793 to
4413.542 ms, whereas isolated construction takes 1.498 to 2.778 ms.
Because bucket capacity and chunk size are equal in this experiment,
larger buckets also yield longer model-execution windows. These windows
provide an opportunity to overlap construction with model computation.
Figure~\ref{fig:maintenance-overlap}(c) shows the median execution-time
difference relative to the model-only baseline. Across configurations,
this difference ranges from 0.674 to 2.277 ms for serial
maintenance and from -0.378 to 0.199 ms with overlap.
Median waiting time for index publication at the next chunk ranges from
2.6 to 3.5 microseconds, indicating that the updated index is available
with little waiting at the point of use.

\FloatBarrier

\section{Conclusion and Future Work}
\label{sec:conclusion}
We presented LongCat Recall Sparse Attention, which represents historical
keys using the centers of capacity-balanced directional groups, retrieves
members by page, and maintains the index incrementally through time
buckets. Capacity constraints
align complete groups with member-page boundaries, allowing fine-grained
representatives to guide regular page access. Fixed page capacities make
the historical refinement workload predictable under a given page budget,
while temporal buckets allow new keys to be indexed without regrouping
existing history. Kernel fusion during index construction and shared
candidate storage reduce kernel launches and repeated data transfers,
respectively. Integration with SGLang extends the design
to asynchronous updates, prefix-based reuse, and index retention, connecting
query-time retrieval with index maintenance and resource management.

Our current retrieval hierarchy uses two stages: center-based page
selection followed by member-level refinement. A deeper hierarchy could
further narrow the search, but adding levels does not necessarily improve
efficiency. At progressively coarser levels, fewer representatives remain
to be evaluated, so another level offers a smaller absolute reduction in
scoring work while still introducing its own scoring and Top-$K$ selection
operations. These additional costs can offset the savings from further
narrowing the search. An important direction for future work is therefore
to design scoring and Top-$K$ kernels that reduce the overhead of traversing
multiple levels, with cross-level fusion as one possible approach.
Such kernels could make deeper retrieval hierarchies more effective at
reducing search cost for longer contexts.

\bibliographystyle{unsrtnat}
\begingroup
\interlinepenalty=10000
\bibliography{references}
\endgroup

\end{document}